\documentclass[11pt]{article}

\usepackage[margin=1in]{geometry}

\usepackage{microtype}
\usepackage{graphicx}
\usepackage{subcaption}
\usepackage{booktabs}

\usepackage{hyperref}
\hypersetup{
    colorlinks=true,
    linkcolor=blue,
    citecolor=blue,
    urlcolor=blue
}

\usepackage{amsmath}
\usepackage{amssymb}
\usepackage{mathtools}
\usepackage{amsthm}

\usepackage[capitalize,noabbrev]{cleveref}

\theoremstyle{plain}

\theoremstyle{definition}

\theoremstyle{remark}

\title{\textbf{Invertible Memory Flow Networks}}

\author{
    \textbf{Liyu Zerihun}\textsuperscript{1} \quad \textbf{Alexandr Plashchinsky}\textsuperscript{1} \\[0.5em]
    \textsuperscript{1}VECTOR Labs \\[0.3em]
    \texttt{liyulg0@gmail.com}, \texttt{aplashch@gmail.com}
}

\date{January 2026}

\begin{document}

\maketitle

\begin{abstract}
Long sequence neural memory remains a challenging problem. RNNs and their variants suffer from vanishing gradients, and Transformers suffer from quadratic scaling. Furthermore, compressing long sequences into a finite fixed representation remains an intractable problem due to the difficult optimization landscape. Invertible Memory Flow Networks (IMFN) make long sequence compression tractable through factorization: instead of learning end-to-end compression, we decompose the problem into pairwise merges using a binary tree of ``sweeper'' modules. Rather than learning to compress long sequences, each sweeper learns a much simpler 2$\rightarrow$1 compression task, achieving O($\log N$) depth with sublinear error accumulation in sequence length. For online inference, we distilled into a constant-cost recurrent student achieving O(1) sequential steps. Empirical results validate IMFN on long MNIST sequences and UCF-101 videos, demonstrating compression of high-dimensional data over long sequences.
\end{abstract}

\section{Introduction}

Despite architectural differences, many modern deep neural network architectures share a fundamental structural core: they can be viewed as dynamical systems evolving a hidden vector across their layers. LSTMs, Transformers, and state space models (SSMs) all fit this lens: there exists a shared vector space where a hidden state evolves, and a decoder that conditions outputs based on this state.

Transformers, for example, can be written as a residual dynamical system \cite{vaswani2017attention}:
\begin{equation}
h_{l+1} = h_l + \text{Block}_l(h_l).
\end{equation}
Here the hidden state is the residual stream and each Transformer block adds to it, contributing information to the shared vector space. Layer normalization complicates this picture slightly, but the residual update intuition still applies \cite{ba2016layernorm}. LSTMs similarly maintain a hidden cell state that is updated and passed along through time \cite{hochreiter1997lstm}. SSMs follow the same pattern, evolving a continuous state through selective updates \cite{gu2022s4}.

These architectures can be unified under a simple dynamical system abstraction:
\begin{equation}
m_{t+1} = m_t + f(m_t, x_t).
\end{equation}
We evolve a state by continuously adding information to it over time, then decode for downstream purposes. The core intuition is that there is a shared ``memory space'' that we interact with and decode from. Under this view, a memory trajectory represents the distinct information it contains.

This is key: the trajectory that memory takes through the vector space \textit{is} information. This perspective converts memory from a storage problem to an invertibility problem. Specifically, can we create a flow of memory through a vector space such that we can invert the trajectory?

Learning such a flow end-to-end is intractable. Transformers face quadratic cost in sequence length, making scaling difficult. Additionally, optimization becomes challenging as the model must learn to attend over a very large number of tokens to compress into a finite state space. RNNs and LSTMs suffer from gradient information becoming saturated over long distances, and the bottleneck becomes the problem \cite{pascanu2013difficulty}. We demonstrate this empirically: a full-attention Transformer and a Mamba model \cite{gu2023mamba} with comparable capacity fail to learn this MNIST sequence reconstruction task, despite having sufficient representational power. The optimization landscape for long-horizon invertible compression is too difficult to navigate directly.

IMFN makes this tractable by exploiting the manifold hypothesis. High-dimensional data often lies on low-dimensional manifolds \cite{bengio2013representation}, meaning that a $2\rightarrow 1$ merge can be approximately invertible for task-relevant information. Rather than learning the full trajectory directly, we decompose the problem into simple pairwise merges arranged in a binary tree, each trained to be locally invertible via a reconstruction loss. The tree structure yields $O(\log N)$ depth, and we show empirically that error accumulates sublinearly in sequence length until memory capacity saturates. For online inference, we distill the tree into a recurrent student that achieves $O(1)$ sequential updates while preserving the invertible flow. 

\begin{figure}[t]
    \centering
    \includegraphics[width=0.7\textwidth]{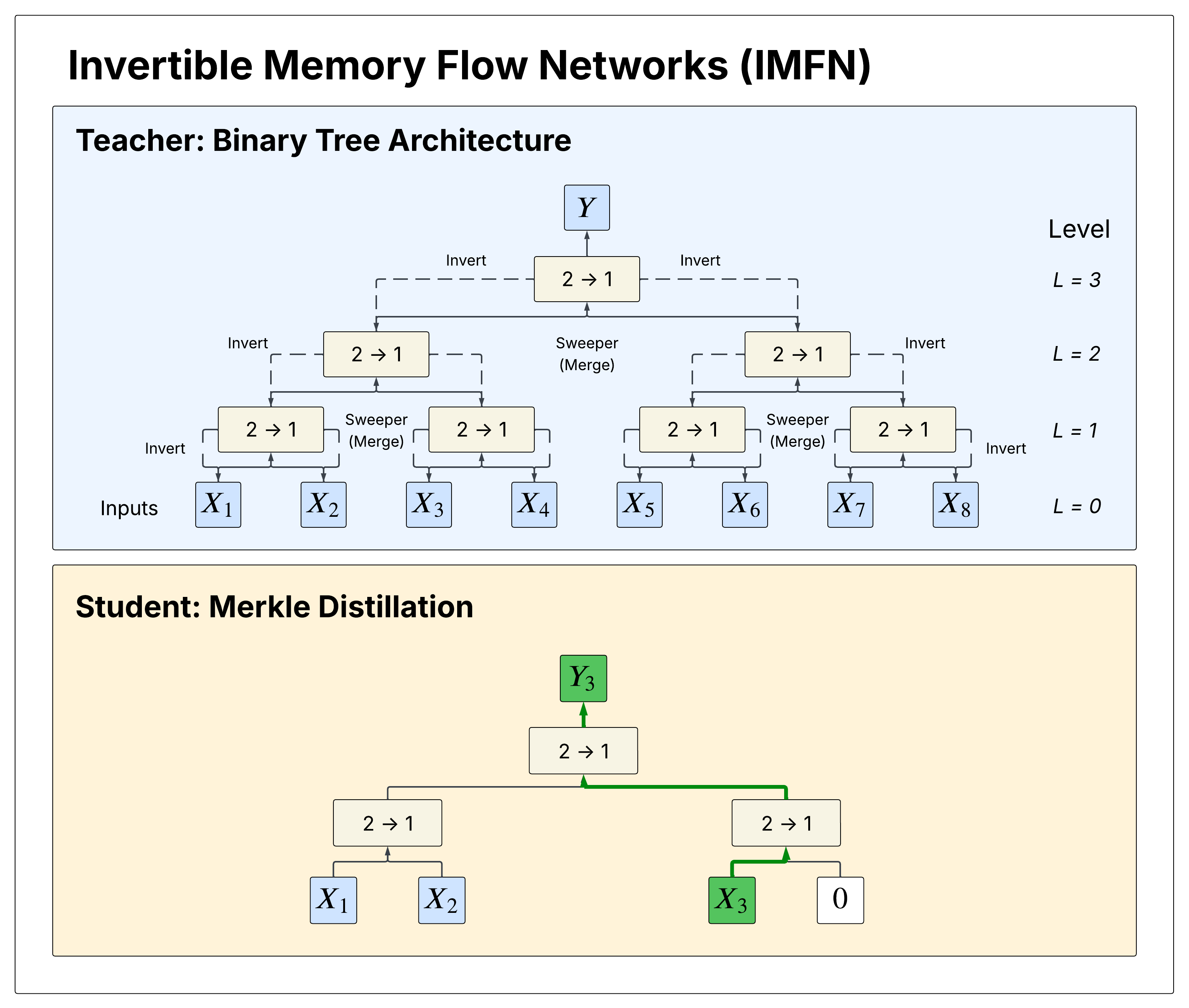}
    \caption{Overview of Invertible Memory Flow Networks (IMFN). 
    The teacher builds a binary-tree memory by repeatedly merging adjacent states with learned ``sweeper'' modules ($2\!\rightarrow\!1$), yielding a single root memory in $O(\log T)$ depth. 
    An inverse pathway applies the corresponding learned decoders ($1\!\rightarrow\!2$) to reconstruct leaf states (and inputs), providing local reconstruction losses that make each merge approximately invertible. 
    For online use, the tree computation is distilled into a recurrent student that updates a fixed-size memory with constant per-step cost. Due to Merkle-style distillation, trajectory simulation using the teacher becomes much more efficient.}
    \label{fig:imfn_overview}
\end{figure}

\section{Related Work}

\subsection{Memory Compression and Recurrent Transformers}

This family of methods extends context by explicitly compressing history. Token pooling and learned downsampling reduce the effective sequence length by aggregating local neighborhoods into fewer summary tokens, enabling attention to scale more favorably with length. Related approaches compress older segments into a smaller memory representation while keeping a short high-resolution window, as in Compressive Transformers \cite{rae2020compressive}. These methods trade exact retention of all past tokens for a compact summary designed to preserve task-relevant information. Recurrent transformer variants aim to maintain long-range dependencies without attending to the entire past at every step. Transformer-XL introduces segment-level recurrence by caching hidden states from previous segments, providing longer effective context without full quadratic recomputation \cite{dai-etal-2019-transformer}. While these approaches improve scalability and empirical long-context behavior, they often rely on learned compression or truncation heuristics that remain globally entangled with the main objective. In contrast, IMFN proposes a structured compression mechanism built from a fixed local primitive (2$\rightarrow$1 merging) applied recursively, yielding logarithmic-depth compression and enabling reconstruction pressure via its paired inversion pathway.

\subsection{Distillation for Online Inference}

Distillation transfers knowledge from a larger or more expensive teacher model into a smaller student model that is cheaper to run at inference time \cite{hinton2015distilling}. In long-context settings, this is especially attractive, as a teacher may perform expensive global computation (e.g., deep recurrence, large attention windows, or structured multi-stage processing), while the student learns to approximate the teacher's behavior using fewer parameters or fewer sequential steps. Distillation has been widely used to compress transformer models while retaining performance \cite{jiao-etal-2020-tinybert}. IMFN uses distillation in a slightly different way: rather than only matching a teacher's outputs, the student is trained to match an \emph{online trajectory} through a fixed-size state space induced by the teacher's tree computation. This connects our approach to a broader line of work on learning predictive state updates that approximate expensive global computation with cheap local transitions.

\subsection{State Space Sequence Models}

Modern SSM-based sequence models seek long-range modeling with linear-time updates by defining a structured latent dynamical system \cite{gu2022s4}. These models emphasize stable long-horizon state propagation and efficient online updates. IMFN is compatible with this goal, but differs in its inductive bias: we construct a compression operator by composing learned local merges in a tree, and we explicitly train an inverse pathway that pressures each merge to preserve information.

\subsection{Other Works}

IMFN is related to bounded-memory sequence models (e.g., RNN/LSTM/GRU) \cite{pascanu2013difficulty, hochreiter1997lstm, cho2014learning}, reversible/invertible architectures \cite{gomez2017reversible, kingma2018glow}, and token pooling/merging methods for efficient attention \cite{ryoo2021tokenlearner, bolya2023tome}. Our focus differs in that we train a fixed local $2\rightarrow1$ merge primitive with an explicit inverse pathway and compose it in a binary tree to enable logarithmic-depth compression and distillation to constant-cost online updates.

\section{Method}

\subsection{IMFN for MNIST Data}

\paragraph{Dataset and preprocessing.}
We use the MNIST training set (60{,}000 images) and treat each image as a single ``frame'' in a longer sequence. Each image is converted to a flattened vector in $[0,1]^{784}$ using a standard tensor transform. For each random seed, we create a fixed 90/10 train--test split, and save the resulting indices to disk to ensure reproducibility across runs.

\paragraph{Teacher architecture.}
The IMFN teacher implements a binary-tree compression mechanism using a stack of level-wise \emph{sweeper} modules. Each sweeper is composed of (i) an encoder that merges two memory vectors into one ($2\!\rightarrow\!1$), and (ii) an inverter that reconstructs two memory vectors from one ($1\!\rightarrow\!2$). For a memory dimension $d$, each sweeper uses a lightweight MLP.

At level $0$, raw pixels are mapped into memory space via an image encoder $E:\mathbb{R}^{784}\rightarrow\mathbb{R}^{d}$ (3-layer MLP with ReLU activations), and reconstructions are produced with an image decoder $D:\mathbb{R}^{d}\rightarrow\mathbb{R}^{784}$ (MLP ending in a sigmoid). For higher levels ($\ell>0$), all merges and reconstructions occur purely in latent space.

\paragraph{Training objective.}
Each sweeper is trained to be locally invertible by minimizing a reconstruction loss between the original pair and the inverted pair after a merge:
\[
(z_L, z_R) \xrightarrow{\text{merge}} \hat{z} \xrightarrow{\text{invert}} (\tilde{z}_L, \tilde{z}_R).
\]
At level $0$, reconstruction is computed in image space:
\[
\mathcal{L}_{\text{recon}}^{(0)} =
\|D(\tilde{z}_L)-x_L\|_2^2 + \|D(\tilde{z}_R)-x_R\|_2^2.
\]
At higher levels, reconstruction is computed directly in latent space:
\[
\mathcal{L}_{\text{recon}}^{(\ell)} =
\|\tilde{z}_L-z_L\|_2^2 + \|\tilde{z}_R-z_R\|_2^2,\quad \ell>0.
\]
To prevent unbounded growth in merged representations, we add a small norm regularizer on the merged code:
\[
\mathcal{L} = \mathcal{L}_{\text{recon}} + \lambda \|\hat{z}\|_2^2,
\]
with $\lambda = 10^{-3}$. During training we also inject small Gaussian noise into the merged code $\hat{z}$ to improve robustness ($\sigma=10^{-2}$).

\paragraph{Level-wise training procedure (factorized training).}
Rather than training end-to-end on full sequences, we train IMFN \emph{factorized by depth}. We maintain a \emph{bank} of representations at each level:
\begin{itemize}
    \item \textbf{Bank 0:} flattened MNIST images $x \in \mathbb{R}^{784}$.
    \item \textbf{Bank $\ell>0$:} merged latents $z^{(\ell)} \in \mathbb{R}^{d}$ produced by the sweeper at level $\ell-1$.
\end{itemize}
To encourage stability under missing information and zero-padding (important for later student distillation), we augment the level-0 bank with 1{,}000 all-zero image vectors and shuffle the combined bank.

Training proceeds sequentially over levels. At each level $\ell$, we optimize only the sweeper at that level using randomly sampled pairs from the corresponding bank: the ``right'' element is taken from the current minibatch, and the ``left'' element is sampled independently as a random partner. After finishing training at level $\ell$, we construct the next bank level $\ell+1$ by repeatedly sampling random pairs from bank $\ell$ and applying the merge function (without inversion) to produce a new latent bank.

\paragraph{Hyperparameters and training grid.}
We train teachers across a grid of 5 random seeds $\{42, 123, 456, 789, 2024\}$ and 5 memory sizes $d \in \{128, 256, 512, 1024, 2048\}$. Optimization uses Adam with learning rate $10^{-4}$ and minibatch size 64. Each level is trained for 50 iterations over the full bank (i.e., 50 epochs per level). For the MNIST teacher, we use 9 sweeper levels in the module list, corresponding to supporting compression depths up to 256 frames (8 merges) while keeping a fixed architecture across experiments.

\paragraph{Evaluation: full-sequence roundtrip reconstruction.}
To measure how well the teacher preserves information across long horizons, we evaluate roundtrip reconstruction on sequences of length $T\in\{16,32,64,128,256\}$ (powers of two). For each $T$, we sample 500 random sequences from the held-out test split. Each sequence is (i) encoded into leaf latents, (ii) merged upward through a binary tree using the first $\log_2 T$ sweepers, and (iii) inverted downward to recover $T$ leaf latents, which are decoded back to pixels. We report mean squared error (MSE) between original and reconstructed images, both (a) averaged across all frame positions and (b) broken down by frame index to analyze positional degradation.

\paragraph{Baseline comparison: Transformer and Mamba sequence compressors.}
To contextualize IMFN against standard end-to-end long-context compressors, we compare to two autoencoding baselines that compress an entire sequence into a single bottleneck vector and then decode back to all frames.

\textbf{Transformer baseline.}
Each frame is embedded with an MLP $\mathbb{R}^{784}\rightarrow\mathbb{R}^{d}$ (3-layer GELU MLP, $d=1024$), a learnable \texttt{[CLS]} token is prepended, and learned positional embeddings are added.
A Transformer encoder (6 layers, 8 heads, feedforward width $4d$, dropout 0.1) produces contextualized tokens, and the \texttt{[CLS]} output is taken as the compressed sequence memory $m \in \mathbb{R}^{1024}$.
A Transformer decoder (6 layers, 8 heads) reconstructs the sequence by cross-attending from learned per-timestep query vectors to $m$, followed by an MLP decoder $\mathbb{R}^{1024}\rightarrow\mathbb{R}^{784}$ applied per frame.

\textbf{Mamba baseline.}
We use the same frame embedding MLP $\mathbb{R}^{784}\rightarrow\mathbb{R}^{1024}$, followed by a Mamba sequence model (12 layers, state size $d_{\text{state}}=16$).
We take the final timestep output as the compressed memory (shape $\mathbb{R}^{1\times 1024}$ to enable cross-attention), and reuse the same Transformer decoder and per-frame MLP decoder as above.
In other words, in this baseline Mamba replaces only the sequence encoder, while decoding is done with the same Transformer decoder for a controlled comparison (see the repository for implementation details).

\textbf{Evaluation protocol.}
We evaluate both baselines at $d=1024$ and sequence length $T=128$.
For this baseline study, we train and evaluate across 4 seeds $\{42,123,456,789\}$ and the same validation split that was used for IMFN teacher.

\subsection{Student Distillation}

The teacher defines a compression function $f(x_1, \ldots, x_n) = y$, where $x_i$ are leaf latents and $y$ is the compressed root memory. To distill this into a model with $O(1)$ sequential updates, we convert the tree computation into a causal relationship.

\paragraph{Trajectory from zeros.} The key idea is to define a trajectory by zero-padding. We set all leaves to zero except those we have ``seen'' so far:
\begin{align}
y_0 &= f(0, 0, \ldots, 0) \\
y_1 &= f(x_1, 0, \ldots, 0) \\
y_2 &= f(x_1, x_2, 0, \ldots, 0) \\
&\vdots \notag \\
y_n &= f(x_1, x_2, \ldots, x_n)
\end{align}
This defines a trajectory $y_0 \rightarrow y_1 \rightarrow \cdots \rightarrow y_n$ through memory space. Each $y_t$ is a well-defined target: the teacher's root when leaves $1, \ldots, t$ contain data and leaves $t+1, \ldots, n$ are zero. The student's task is to learn the transition $y_{t-1} \rightarrow y_t$ given input $x_t$.

\paragraph{Efficient trajectory generation.} A naive approach would rebuild the entire tree at each timestep, requiring $O(n)$ merges per step and $O(n^2)$ total. We observe that the tree structure admits a Merkle-style optimization: when leaf $t$ changes from zero to $x_t$, only the nodes on the path from that leaf to the root are affected. This path has length $\log n$, so each update requires $O(\log n)$ merges. The full trajectory of $n$ teacher targets can thus be generated in $O(n \log n)$ time.

\paragraph{Student architecture.} The student is a feedforward network that predicts the memory delta at each step:
\begin{equation}
m_{t+1} = m_t + g_\theta(m_t, x_t, t)
\end{equation}
where $g_\theta$ is a 4-layer MLP with hidden dimension $2d$ (where $d$ is the memory dimension). The input concatenates the current memory $m_t \in \mathbb{R}^d$, the new latent $x_t \in \mathbb{R}^d$, and a one-hot positional encoding $e_t \in \mathbb{R}^T$. The network outputs a delta $\Delta_t \in \mathbb{R}^d$ that is added to the current memory. This residual formulation mirrors the additive dynamics discussed in Section 1, and ensures the student can represent the identity update when no change is needed.

The explicit positional encoding means the student is not strictly Markovian; however, the critical property is preserved: each update requires $O(1)$ computation and $O(d)$ memory, independent of sequence length. At inference, the student processes a stream of inputs with constant cost per step, compared to the teacher's $O(\log n)$ cost for incremental tree updates.

\paragraph{Training on rollouts.} We train the student on its own rollouts to ensure robustness to the states it will encounter at inference. For each training trajectory:
\begin{enumerate}
    \item Sample $n$ images and compute their latents $x_1, \ldots, x_n$ using the frozen teacher encoder.
    \item Generate teacher targets $y_1, \ldots, y_n$ via the Merkle-style incremental tree construction.
    \item Roll out the student: starting from $m_0 = y_0$ (the all-zeros root), compute $m_1, \ldots, m_n$ by applying the student recurrently. These rollout states are detached from the computation graph.
    \item Sample a subset of timesteps (25\% by default) and compute the loss:
    \begin{equation}
        \mathcal{L} = \frac{1}{|S|} \sum_{t \in S} \| g_\theta(m_{t-1}, x_t, t) + m_{t-1} - y_t \|^2
    \end{equation}
    where $S$ is the sampled subset and $y_t$ is the teacher target.
\end{enumerate}
Training from the student's own states (rather than teacher states) prevents distribution shift at inference, where errors may accumulate and push the student into regions of state space not seen during training.

\paragraph{Experimental setup.} We train students across 5 seeds $\{42, 123, 456, 789, 2024\}$ and sequence lengths $T \in \{16, 32, 64, 128, 256\}$. Each configuration is trained for 1000 epochs with 100 trajectories per epoch, using Adam with learning rate $10^{-4}$. The teacher for each seed is frozen during student training. Train and test splits (90\%/10\%) are fixed across all experiments using a separate split seed to ensure comparable evaluation.

\subsection{IMFN for UCF-101 Videos}

We validate IMFN on real video using UCF-101, using the same teacher-style roundtrip reconstruction protocol as in MNIST but with a tokenized video representation.
To keep the setup easy to reproduce, we use a single fixed seed and provide a complete training/evaluation recipe.

\paragraph{Video preprocessing.}
For each video, we sample a contiguous clip of $T=128$ frames using a deterministic loader (seeded by the experiment seed), resize frames to $64\times 64$, and normalize pixels to $[0,1]$.

\paragraph{Level-0 (L0) representation: pairwise frame tokenizer.}
At the bottom of the hierarchy, L0 merges two RGB frames into a fixed set of memory tokens using a Perceiver-style cross-attention stack.
Each frame is patchified into $8\times 8$ patches (64 patches per frame), embedded into $d=256$ dimensions, concatenated across the two frames (128 input tokens), and compressed into $n_{\text{mem}}=96$ learned memory tokens (memory ratio $0.75$).
The L0 inverter reconstructs the two frames by decoding the memory tokens back into patch tokens and unpatchifying.
Thus, each L0 latent represents a $2{:}1$ temporal compression.

\paragraph{Higher levels (L1+): token merge sweepers.}
Each higher-level sweeper merges two token-latents into one and inverts one latent back into two, operating purely in token space.
Concretely, for $\ell\ge 1$ the merge maps $\mathbb{R}^{(2 n_{\text{mem}})\times d}\rightarrow\mathbb{R}^{n_{\text{mem}}\times d}$ and the inverter maps $\mathbb{R}^{n_{\text{mem}}\times d}\rightarrow\mathbb{R}^{(2 n_{\text{mem}})\times d}$.
Using levels L0--L3 yields effective temporal compression ratios of $2{:}1$ (L0), $4{:}1$ (L1), $8{:}1$ (L2), and $16{:}1$ (L3).

\paragraph{Training objective.}
We train each sweeper with a local token-space reconstruction loss, encouraging $\text{invert}(\text{merge}(z_L, z_R)) \approx (z_L, z_R)$.
To better align higher-level merges with perceptual fidelity, we additionally include a downstream pixel reconstruction loss by decoding a sampled branch through the frozen lower stack back to frames (``path-sampled'' reconstruction), plus a small merged-memory norm penalty.

\paragraph{Evaluation metrics and Reproducibility}
We evaluate each level by encoding validation clips up to that level and decoding back to frames, then report per-clip MSE and PSNR (in dB) averaged over the full validation set.

All UCF-101 results use a fixed train/validation split included with the code, and we provide checkpoints for the bottom of the hierarchy (L0 and L1).
Higher levels (L2 and L3) can be retrained deterministically using the same level-wise procedure and the released lower-level checkpoints.

\section{Results}

\subsection{IMFN Teacher on MNIST Sequences}

We first evaluate the IMFN teacher on MNIST sequences under full roundtrip compression and decompression (Section 3.1). Across all IMFN specific experiments, results are averaged over 5 random seeds $\{42,123,456,789,2024\}$, and reconstruction quality is measured using mean squared error (MSE) between original and decoded frames. For the baseline comparison we used 4 seeds $\{42,123,456,789\}$. 

\begin{figure}[t]
    \centering
    \includegraphics[width=0.7\textwidth]{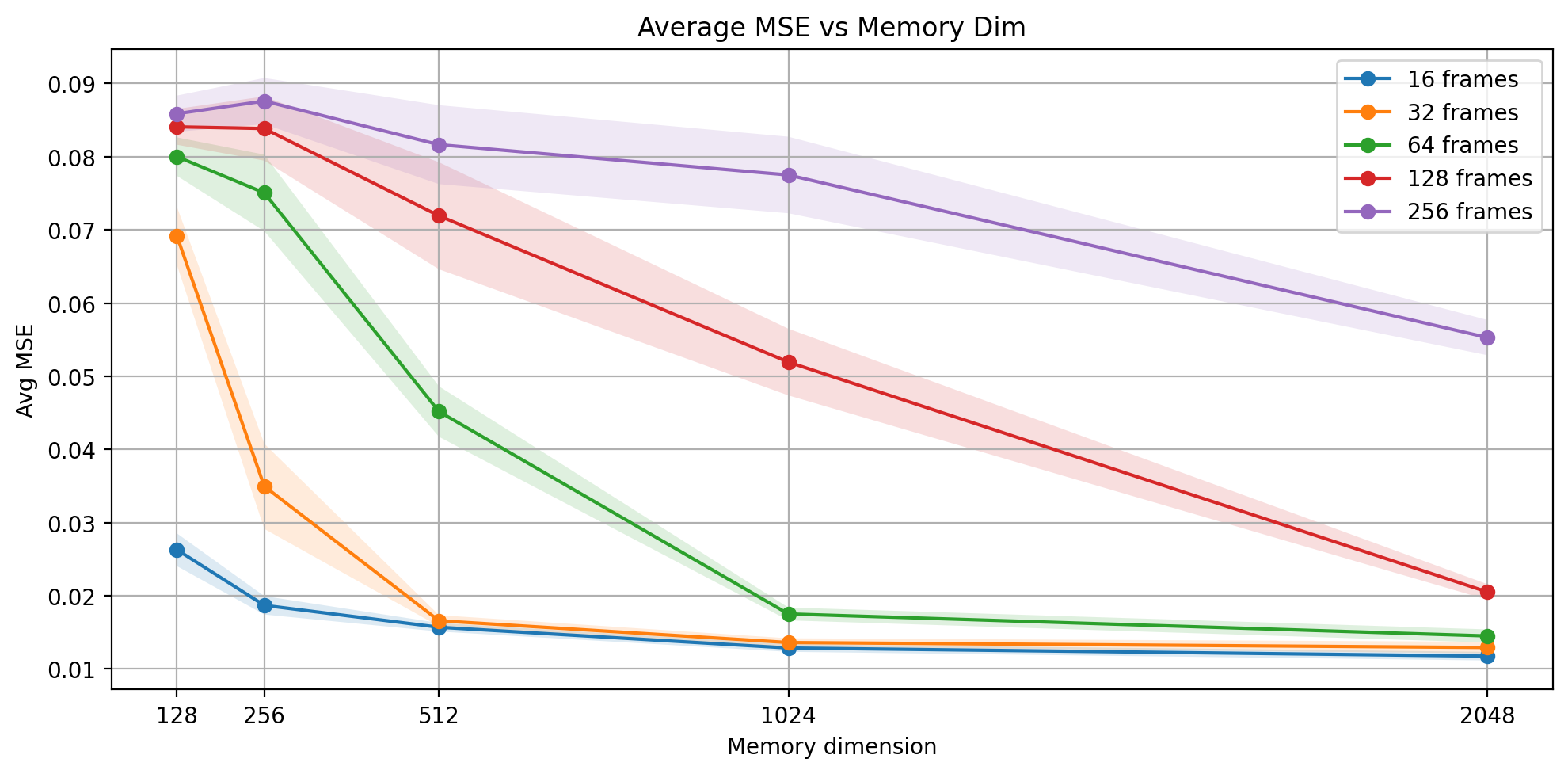}
    \caption{IMFN teacher reconstruction error (MNIST). Average MSE vs.\ memory dimension $d$, averaged over 5 seeds.}
    \label{fig:mnist_mse_vs_dim}
\end{figure}

\paragraph{Effect of memory dimension.}
Figure~\ref{fig:mnist_mse_vs_dim} shows average reconstruction MSE as a function of memory dimension $d \in \{128,256,512,1024,2048\}$, with separate curves for sequence lengths $T \in \{16,32,64,128,256\}$. Increasing memory capacity consistently improves reconstruction fidelity, with the effect becoming substantially stronger at longer horizons. For short sequences ($T=16$ and $T=32$), performance saturates quickly: modest memory sizes already achieve low MSE and additional capacity yields diminishing returns. In contrast, longer sequences ($T=128$ and $T=256$) exhibit a pronounced dependency on memory dimension, indicating that reconstruction quality is bottlenecked primarily by representational capacity rather than optimization instability at these lengths. Notably, for the longest sequences, increasing $d$ continues to reduce MSE even up to $d=2048$, suggesting that the teacher has not yet reached a hard saturation point in this regime.

\begin{figure}[t]
    \centering
    \includegraphics[width=0.7\textwidth]{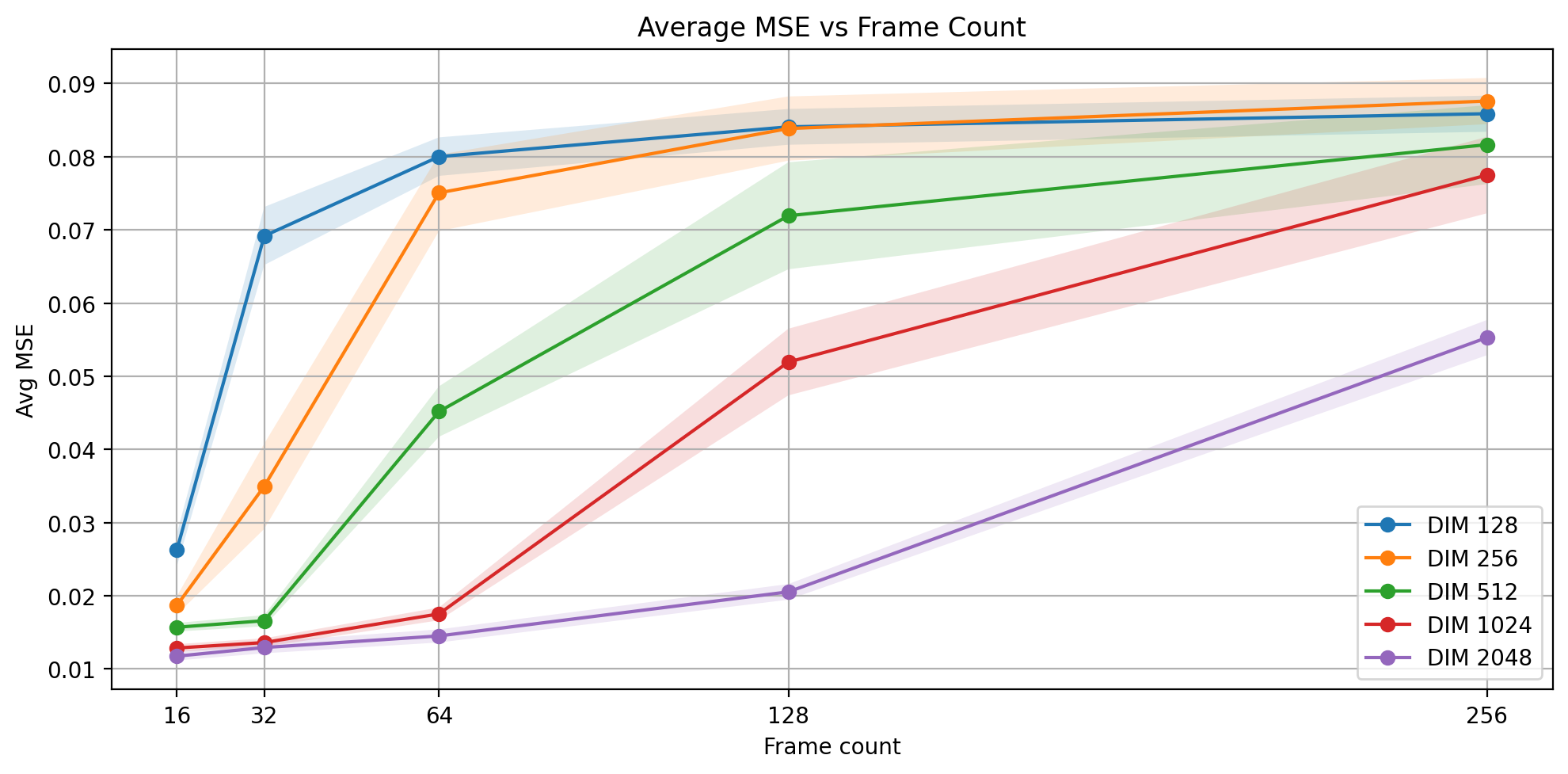}
    \caption{IMFN teacher reconstruction error scaling with horizon. Average MSE vs.\ sequence length $T$, averaged over 5 seeds.}
    \label{fig:mnist_mse_vs_frames}
\end{figure}

\paragraph{Effect of sequence length and error accumulation.}
Figure~\ref{fig:mnist_mse_vs_frames} summarizes how reconstruction MSE scales as sequence length increases, shown separately for each memory dimension. For all $d$, longer sequences increase reconstruction error, reflecting accumulation of information loss across repeated merges and inversions. However, the rate of degradation strongly depends on memory capacity: larger $d$ reduces the slope of this degradation, preserving reconstruction quality over substantially longer horizons. In particular, the gap between short and long sequences shrinks as memory dimension increases, consistent with IMFN's hypothesis that the local $2\!\rightarrow\!1$ merge primitive becomes effectively invertible when sufficient capacity is available for the merged representation.

\subsection{IMFN Student on MNIST Sequences}
\label{sec:mnist_student}

We evaluate the distilled student that updates a single memory vector online while processing a stream of frames.
Unless stated otherwise, we report pixel-space MSE on held-out sequences, averaged over 5 random seeds.
For each horizon $T \in \{16,32,64,128,256\}$ we train a separate student (the student uses a $T$-length positional encoding), and teacher targets at step $t$ depend on $T$ because unseen leaves are zero-padded.
To make comparisons consistent, both student and teacher states are decoded using the same frozen teacher inversion stack and image decoder.

\begin{figure}[t]
    \centering
    \includegraphics[width=0.7\textwidth]{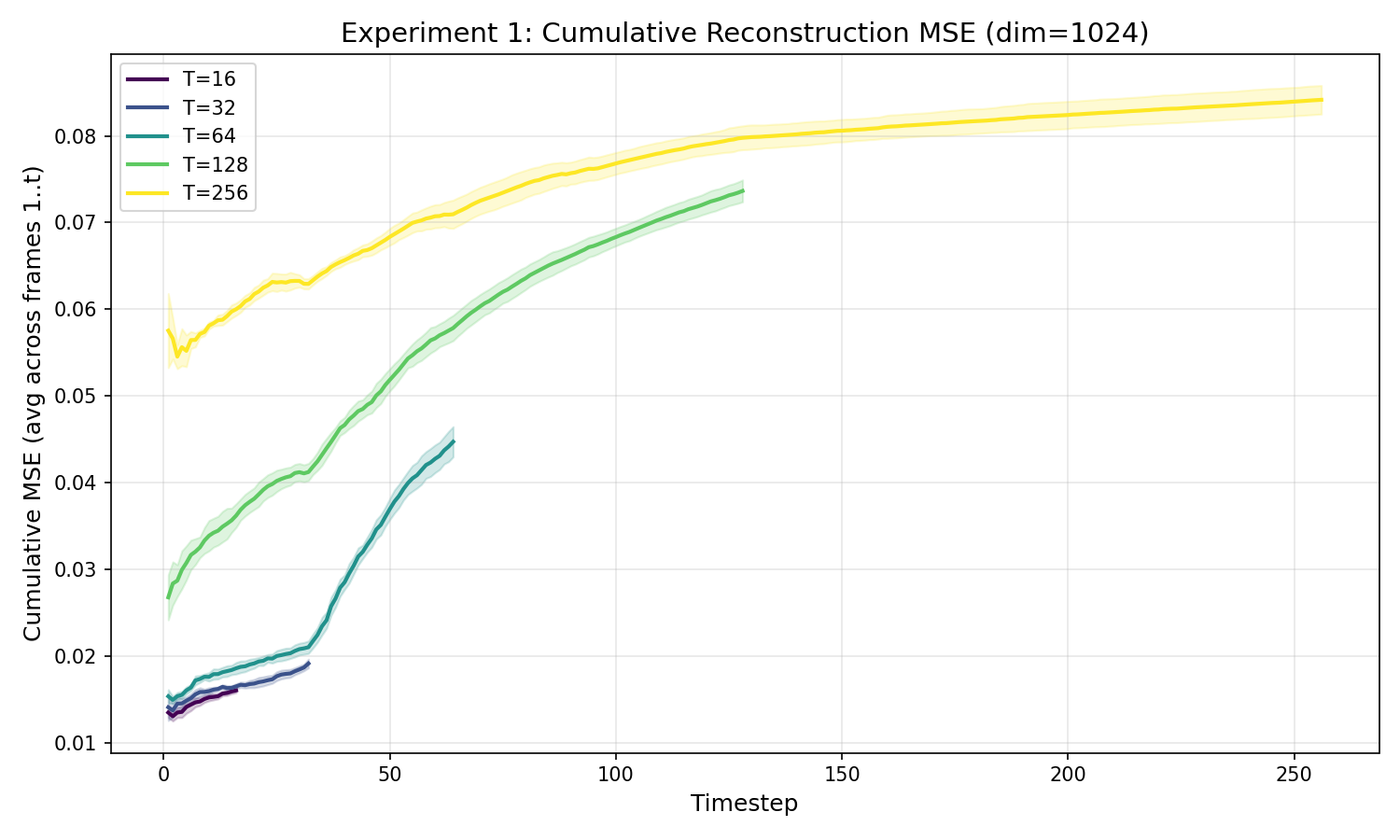}
    \caption{
    Student online prefix reconstruction (MNIST, $d=1024$).
    After $t$ online updates, we decode the first $t$ frames and report MSE averaged over the prefix.
    Curves correspond to different horizons $T$ (separate student per $T$).
    }
    \label{fig:student_prefix_mse}
\end{figure}

\begin{figure}[t]
    \centering
    \includegraphics[width=0.7\textwidth]{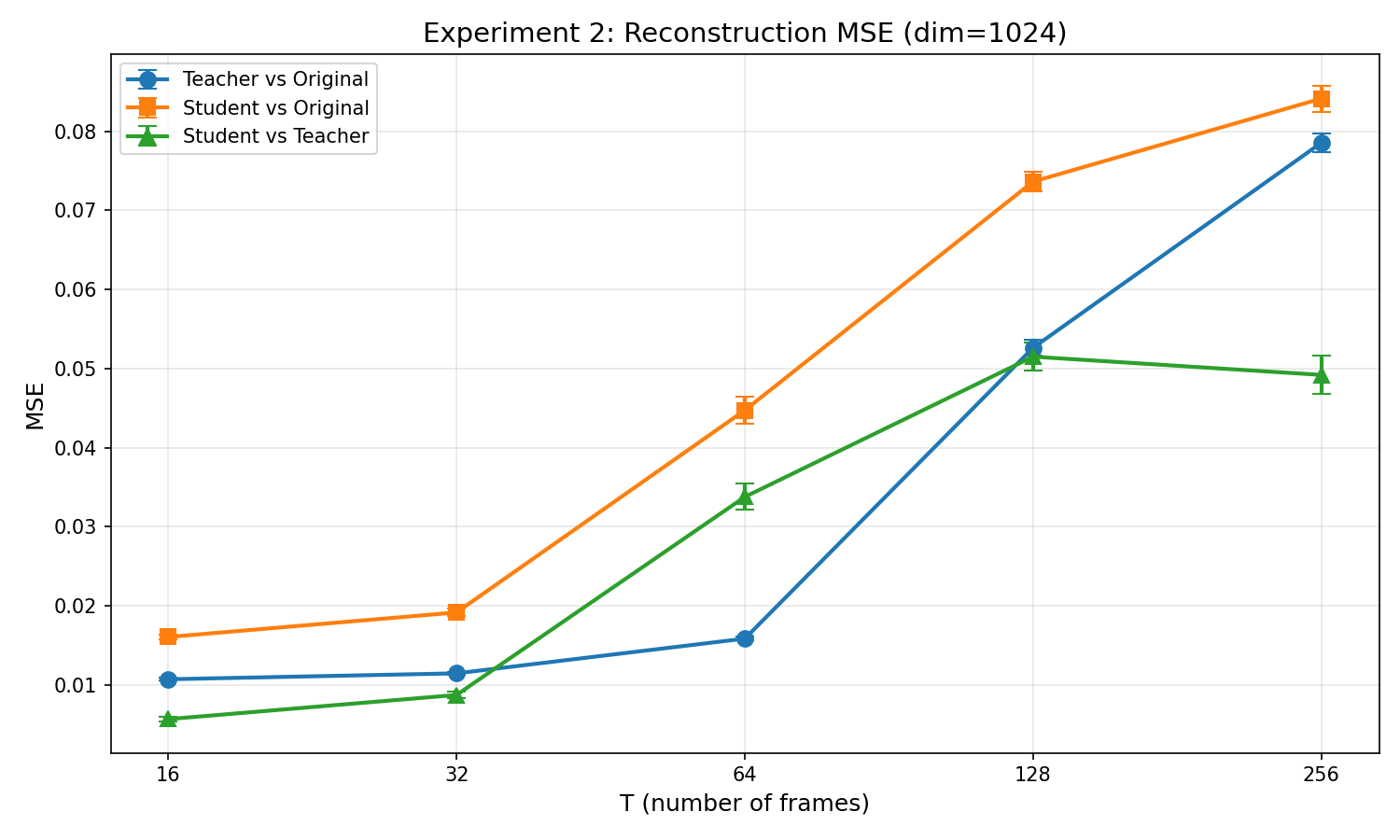}
    \caption{
    End-of-sequence reconstruction scaling with horizon (MNIST, $d=1024$).
    Pixel MSE of teacher and student reconstructions vs.\ the original sequence at the final step.
    Error bars show variance across seeds.
    }
    \label{fig:student_mse_vs_T}
\end{figure}

Figure~\ref{fig:student_prefix_mse} measures \emph{online prefix retention}.
After $t$ online updates, we decode the first $t$ reconstructed frames and compute MSE averaged over the prefix.
Reconstruction error increases with $t$, and the increase is steeper for larger horizons. We summarize end-of-sequence performance in Figure~\ref{fig:student_mse_vs_T}.

\subsection{MNIST Sequence Compression: IMFN vs Transformer vs Mamba}

We compare IMFN against two end-to-end baselines that compress an entire sequence into a single bottleneck vector: (i) a full-attention Transformer compressor and (ii) a Mamba-based compressor (Section~3.1).
All three models are evaluated on the same MNIST sequence reconstruction task with sequence length $T=128$ and memory dimension $d=1024$, across 4 random seeds $\{42,123,456,789\}$ (500 held-out test sequences per seed). Table~\ref{tab:mnist_imfn_vs_transformer_vs_mamba_128} reports mean $\pm$ std across seeds.
IMFN achieves the best overall reconstruction quality, with Mamba consistently outperforming the Transformer baseline (notably in SSIM).

\begin{table}[t]
\centering
\begin{tabular}{lccc}
\toprule
Model & MSE ($\downarrow$) & PSNR (dB) ($\uparrow$) & SSIM ($\uparrow$) \\
\midrule
IMFN        & $0.052132 \pm 0.000939$ & $12.83 \pm 0.08$ & $0.6300 \pm 0.0058$ \\
Transformer & $0.066598 \pm 0.000140$ & $11.77 \pm 0.01$ & $0.3396 \pm 0.0011$ \\
Mamba       & $0.055753 \pm 0.000518$ & $12.54 \pm 0.04$ & $0.4341 \pm 0.0041$ \\
\bottomrule
\end{tabular}
\caption{MNIST sequence compression ($T=128$, $d=1024$). Mean $\pm$ std across seeds $\{42,123,456,789\}$.}
\label{tab:mnist_imfn_vs_transformer_vs_mamba_128}
\end{table}

\begin{table}[t]
\centering
\begin{tabular}{lrr}
\toprule
Model & Parameters & Raw \\
\midrule
IMFN & 61.43M & 61{,}425{,}424 \\
Transformer & 182.56M & 182{,}558{,}480 \\
Mamba & 186.86M & 186{,}855{,}184 \\
\bottomrule
\end{tabular}
\caption{Model sizes for the MNIST sequence compression experiment ($T=128$, $d=1024$).}
\label{tab:model_sizes}
\end{table}

\begin{figure}[t]
    \centering
    \includegraphics[width=0.9\textwidth]{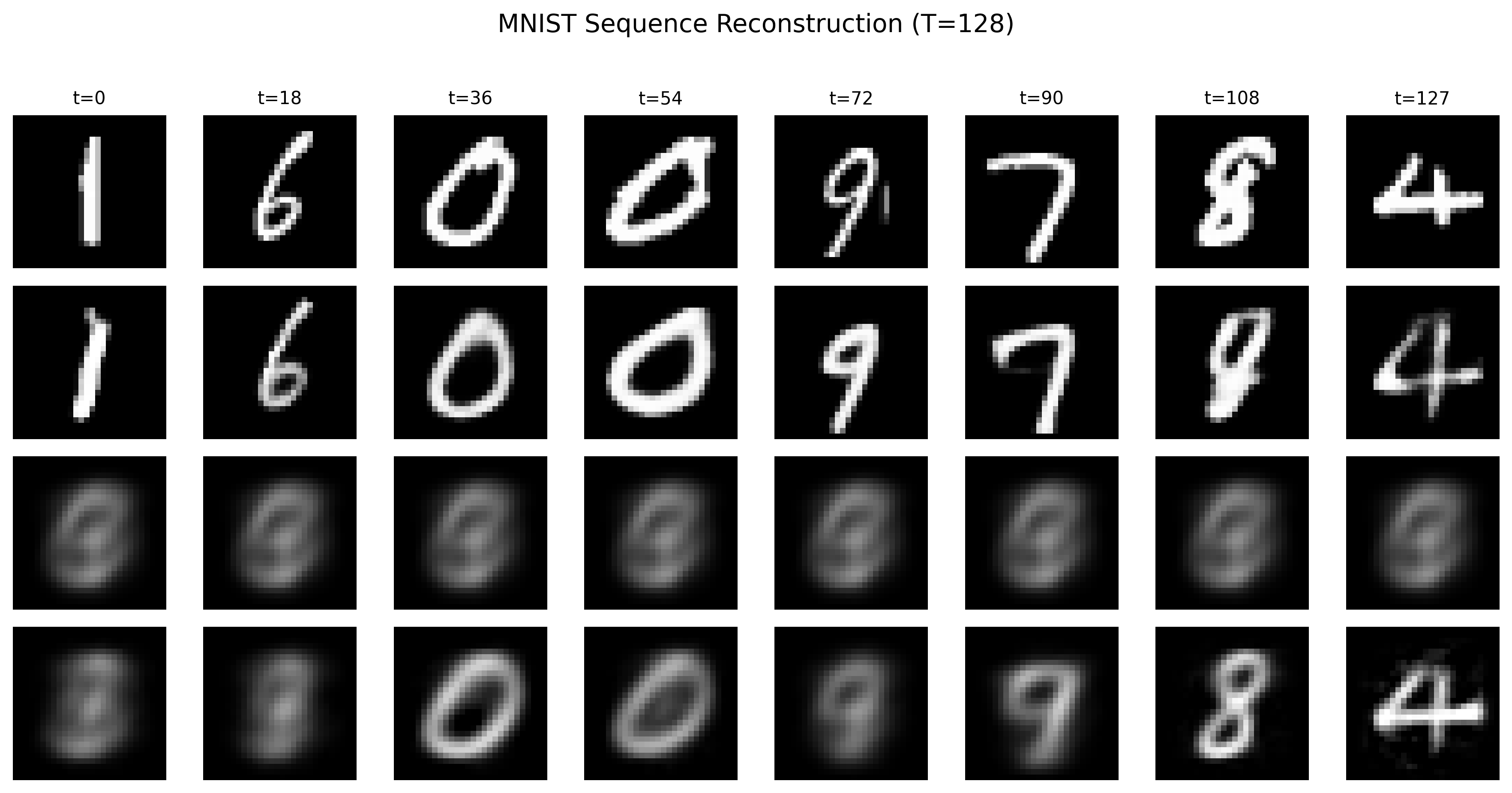}
    \caption{
    MNIST sequence compression at $T=128$ and $d=1024$.
    Top: original frames.
    Bottom rows: reconstructions after compression--decompression with IMFN, Transformer, and Mamba.
    }
    \label{fig:mnist_imfn_vs_transformer_vs_mamba_128}
\end{figure}

\paragraph{Qualitative reconstruction fidelity.}
Figure~\ref{fig:mnist_imfn_vs_transformer_vs_mamba_128} shows a representative visualization where the top row is the original sequence and subsequent rows are reconstructions after compression--decompression.
IMFN preserves digit identity and stroke structure most reliably.
The Transformer baseline tends to blur or collapse digits toward ``prototype-like'' shapes, while Mamba improves detail retention relative to Transformer but still lags IMFN.

Even when average MSE values are close, the reconstructions can look very different: MSE is a purely pixel-wise metric, so it can assign similar scores to images where the error is spread out as mild blur versus images where the key edges and strokes are preserved but a few pixels are off. In our case, a simple hypothesis is that IMFN's approximate invertibility encourages each merge to retain information that can be recovered by the inverse path, which tends to preserve semantic structure (digit identity) rather than drifting toward ``average'' prototypes. This is consistent with SSIM: IMFN scores substantially higher and its reconstructions remain clean and recognizable as the correct class across frames.

\paragraph{The optimization gap (end-to-end compression).}
Both end-to-end baselines have sufficient representational capacity in principle (same bottleneck dimension $d=1024$) but must learn a single global mapping from all frames to a fixed vector.
IMFN avoids this global optimization problem by factorizing compression into locally trained $2\!\rightarrow\!1$ merges with an explicit inversion pathway, yielding more stable long-horizon information preservation.

\subsection{UCF-101 Video Compression Results}

We evaluate UCF-101 reconstructions on the full validation set (1{,}332 videos), reporting per-video MSE and PSNR after encoding to a given level and decoding back to frames.
Table~\ref{tab:ucf_validation_mse_psnr} shows that reconstruction quality degrades gracefully as the temporal compression ratio increases from $2{:}1$ (L0) to $16{:}1$ (L3).

\begin{table}[t]
\centering
\begin{tabular}{lcc}
\toprule
Level & MSE ($\downarrow$) & PSNR (dB) ($\uparrow$) \\
\midrule
L0 (2{:}1)  & $0.000796 \pm 0.000551$ & $31.95 \pm 2.96$ \\
L1 (4{:}1)  & $0.000810 \pm 0.000556$ & $31.87 \pm 2.96$ \\
L2 (8{:}1)  & $0.001204 \pm 0.000888$ & $30.25 \pm 3.11$ \\
L3 (16{:}1) & $0.002163 \pm 0.001839$ & $27.96 \pm 3.46$ \\
\bottomrule
\end{tabular}
\caption{UCF-101 validation reconstruction performance (mean $\pm$ std over 1{,}332 videos) at different compression depths.}
\label{tab:ucf_validation_mse_psnr}
\end{table}

\paragraph{Sublinear error growth with compression ratio.}
Each higher level doubles the temporal compression ratio (2{:}1 $\rightarrow$ 4{:}1 $\rightarrow$ 8{:}1 $\rightarrow$ 16{:}1).
Across these $8\times$ harder compression settings (L0 to L3), average reconstruction MSE increases by only $\approx 2.7\times$ ($7.96\times 10^{-4}$ to $2.16\times 10^{-3}$), i.e. error grows sublinearly with compression ratio.
A linear-in-ratio baseline would predict roughly $0.000796 \times 8 \approx 0.0064$ MSE at 16{:}1, substantially worse than the observed $0.0022$.

\section{Discussion}

\paragraph{Parameter efficiency via invertibility.}
A key finding of this work is that IMFN outperforms baselines that are about 3$\times$ larger (61M vs.\ $\sim$180M parameters). This result challenges the prevailing assumption that scale is the primary driver of performance. We attribute this efficiency to the factorization of the optimization problem: compressing 128 high-dimensional images into a single vector creates a difficult optimization landscape for end-to-end models. By enforcing local invertibility at every step, IMFN turns this global problem into a sequence of manageable local problems. Remarkably, our best models compress MNIST sequences down to just 8 scalars per image (memory dimension $d=1024$ distributed over $T=128$ frames), yet retain enough fidelity to reconstruct digits clearly. This suggests that when the optimization is structured correctly, strong compression can be achieved without simply scaling parameter count.

\paragraph{Structural memory and the Merkle prior.}
Standard sequence models tend to treat memory as either a flat context window (Transformers) or a streaming state (RNNs and SSMs). IMFN introduces a structural prior: memory as a binary tree. This structure is not just an implementation detail but a functional advantage. It allows ``Merkle-style'' updates, where changing a single leaf only requires $O(\log N)$ recomputations to update the root memory. This connection between data structures (Merkle trees) and differentiable memory suggests a path toward neural memory systems that are both semantically rich and efficient to update.

\paragraph{The student teacher gap.}
While the student model achieves $O(1)$ inference, performance degrades at longer sequence lengths compared to the teacher. One likely contributor is training schedule: we trained students for the same number of epochs regardless of sequence length $T$. Modeling the trajectory induced by a deeper tree is substantially harder than modeling a shallow one. Closing this gap may require longer training schedules and intermediate supervision, for example by forcing the student to match intermediate partial-tree states in addition to the final root state.

\paragraph{Benchmarking autoencoding baselines and harder benchmarks}
These baselines (Mamba and Transformer) are often optimized for next-token prediction, where discarding high-frequency detail can even be beneficial. Our results do not claim IMFN is superior for generative modeling in general, but rather that it offers an advantage on tasks that require high-fidelity compression and storage. 

While MNIST sequences provide a clean testbed, stronger evidence will require benchmarking compression at longer horizons and under more realistic modalities, where information must be retained over hundreds to thousands of steps. We also view broader comparisons to modern long-context methods as essential.

\section{Conclusion}

We presented Invertible Memory Flow Networks (IMFN), a framework that treats neural memory not as a static container, but as a trajectory through a vector space. By decomposing long-sequence compression into a hierarchy of locally invertible merges, we make the optimization landscape tractable and outperform substantially larger baselines on reconstruction tasks. By distilling this hierarchy into a recurrent student, we also show that it is possible to combine the representational power of deep tree structures with constant-time inference.

Our results serve as evidence that invertibility can be a practical and parameter-efficient route to long-context memory. While end-to-end learning remains dominant, we believe structural decomposition, in particular building invertible flows, is a useful alternative when the goal is stable information retention over long horizons.

\bibliography{references}

@inproceedings{pascanu2013difficulty,
  title={On the difficulty of training recurrent neural networks},
  author={Pascanu, Razvan and Mikolov, Tomas and Bengio, Yoshua},
  booktitle={Proceedings of the 30th International Conference on Machine Learning (ICML)},
  pages     = {1310--1318},
  year={2013}
}

@article{hochreiter1997lstm,
  title   = {Long Short-Term Memory},
  author  = {Hochreiter, Sepp and Schmidhuber, J{\"u}rgen},
  journal = {Neural Computation},
  volume  = {9},
  number  = {8},
  pages   = {1735--1780},
  year    = {1997}
}

@inproceedings{cho2014learning,
  title     = {Learning Phrase Representations using {RNN} Encoder--Decoder for Statistical Machine Translation},
  author    = {Cho, Kyunghyun and van Merri{\"e}nboer, Bart and Gulcehre, Caglar and Bahdanau, Dzmitry and Bougares, Fethi and Schwenk, Holger and Bengio, Yoshua},
  booktitle = {Proceedings of the 2014 Conference on Empirical Methods in Natural Language Processing ({EMNLP})},
  pages     = {1724--1734},
  year      = {2014}
}

@inproceedings{rae2020compressive,
  title     = {Compressive Transformers for Long-Range Sequence Modelling},
  author    = {Rae, Jack W. and Potapenko, Anna and Jayakumar, Siddhant M. and Hillier, Chloe and Lillicrap, Timothy P.},
  booktitle = {International Conference on Learning Representations (ICLR)},
  year      = {2020},
  url       = {https://openreview.net/forum?id=SylKikSYDH}
}

@inproceedings{dai-etal-2019-transformer,
  title     = {Transformer-{XL}: Attentive Language Models beyond a Fixed-Length Context},
  author    = {Dai, Zihang and Yang, Zhilin and Yang, Yiming and Carbonell, Jaime and Le, Quoc and Salakhutdinov, Ruslan},
  booktitle = {Proceedings of the 57th Annual Meeting of the Association for Computational Linguistics},
  pages     = {2978--2988},
  year      = {2019},
  publisher = {Association for Computational Linguistics},
  url       = {https://aclanthology.org/P19-1285/},
  doi       = {10.18653/v1/P19-1285}
}

@article{hinton2015distilling,
  title  = {Distilling the Knowledge in a Neural Network},
  author = {Hinton, Geoffrey and Vinyals, Oriol and Dean, Jeff},
  journal = {arXiv preprint arXiv:1503.02531},
  year   = {2015},
  url    = {https://arxiv.org/abs/1503.02531}
}

@inproceedings{jiao-etal-2020-tinybert,
  title     = "{T}iny{BERT}: Distilling {BERT} for Natural Language Understanding",
  author    = "Jiao, Xiaoqi  and
               Yin, Yichun  and
               Shang, Lifeng  and
               Jiang, Xin  and
               Chen, Xiao  and
               Li, Linlin  and
               Wang, Fang  and
               Liu, Qun",
  booktitle = "Findings of the Association for Computational Linguistics: EMNLP 2020",
  pages     = "4163--4174",
  year      = "2020",
  address   = "Online",
  publisher = "Association for Computational Linguistics",
  url       = "https://aclanthology.org/2020.findings-emnlp.372/",
  doi       = "10.18653/v1/2020.findings-emnlp.372"
}

@inproceedings{vaswani2017attention,
  title     = {Attention Is All You Need},
  author    = {Vaswani, Ashish and Shazeer, Noam and Parmar, Niki and Uszkoreit, Jakob and Jones, Llion and Gomez, Aidan N. and Kaiser, Lukasz and Polosukhin, Illia},
  booktitle = {Advances in Neural Information Processing Systems (NeurIPS)},
  year      = {2017}
}

@article{ba2016layernorm,
  title   = {Layer Normalization},
  author  = {Ba, Jimmy Lei and Kiros, Jamie Ryan and Hinton, Geoffrey E.},
  journal = {arXiv preprint arXiv:1607.06450},
  year    = {2016}
}

@article{bengio2013representation,
  title   = {Representation Learning: A Review and New Perspectives},
  author  = {Bengio, Yoshua and Courville, Aaron and Vincent, Pascal},
  journal = {IEEE Transactions on Pattern Analysis and Machine Intelligence},
  volume  = {35},
  number  = {8},
  pages   = {1798--1828},
  year    = {2013}
}

@inproceedings{gu2022s4,
  title     = {Efficiently Modeling Long Sequences with Structured State Spaces},
  author    = {Gu, Albert and Goel, Karan and R{\'e}, Christopher},
  booktitle = {International Conference on Learning Representations (ICLR)},
  year      = {2022}
}

@inproceedings{gomez2017reversible,
  title     = {The Reversible Residual Network: Backpropagation Without Storing Activations},
  author    = {Gomez, Aidan N. and Ren, Mengye and Urtasun, Raquel and Grosse, Roger B.},
  booktitle = {Advances in Neural Information Processing Systems (NeurIPS)},
  pages   = {2214--2224},
  year      = {2017}
}

@inproceedings{kingma2018glow,
  title     = {Glow: Generative Flow with Invertible $1\times1$ Convolutions},
  author    = {Kingma, Diederik P. and Dhariwal, Prafulla},
  booktitle = {Advances in Neural Information Processing Systems (NeurIPS)},
  pages   = {10236--10245},
  year      = {2018}
}

@inproceedings{ryoo2021tokenlearner,
  title     = {TokenLearner: Adaptive Space-Time Tokenization for Videos},
  author    = {Ryoo, Michael S. and Piergiovanni, A. J. and Arnab, Anurag and Dehghghani, Mostafa and Angelova, Anelia},
  booktitle = {Advances in Neural Information Processing Systems},
  volume    = {34},
  pages     = {12786--12797},
  year      = {2021},
  url       = {https://proceedings.neurips.cc/paper_files/paper/2021/hash/6a30e32e56fce5cf381895dfe6ca7b6f-Abstract.html}
}

@inproceedings{bolya2023tome,
  title     = {Token Merging: Your {ViT} but Faster},
  author    = {Bolya, Daniel and Fu, Cheng-Yang and Dai, Xiaoliang and Zhang, Peizhao and Feichtenhofer, Christoph and Hoffman, Judy},
  booktitle = {International Conference on Learning Representations},
  year      = {2023},
  url       = {https://openreview.net/forum?id=JroZRaRw7Eu}
}

@article{gu2023mamba,
  title   = {Mamba: Linear-Time Sequence Modeling with Selective State Spaces},
  author  = {Gu, Albert and Dao, Tri},
  journal = {arXiv preprint arXiv:2312.00752},
  year    = {2023}
}
\bibliographystyle{plain}

\appendix

\section{Experimental Details and Reproducibility}
\label{app:repro}

\subsection{MNIST sequences: data and protocol}
We use the MNIST training set (60{,}000 images) and create a fixed 90/10 train--test split per random seed. Each image is flattened to $x\in[0,1]^{784}$. Sequence datasets are constructed by sampling $T\in\{16,32,64,128,256\}$ images per sequence.

\subsection{MNIST teacher training (IMFN)}
We train the IMFN teacher level-by-level. Each level is trained to be locally invertible via a reconstruction loss, with level 0 computed in pixel space and higher levels computed in latent space. We add a small merged-code norm penalty ($\lambda=10^{-3}$) and inject small Gaussian noise into the merged code during training ($\sigma=10^{-2}$).

\subsection{UCF-101 videos: preprocessing and deterministic clips}
For UCF-101, we decode videos to frames, resize to $64\times 64$, and normalize pixels to $[0,1]$. For reproducibility, each video uses a deterministic clip window based on a stable hash of (seed, video path, sequence length), and short videos are padded by repeating the last frame to ensure a fixed $T$.

\subsection{UCF-101 L0 and higher-level sweepers}
At level 0, each pair of RGB frames is patchified into $8\times 8$ patches and embedded into $d=256$ dimensions. The two frames are concatenated into 128 input tokens and compressed into $n_{\mathrm{mem}}=96$ memory tokens (memory ratio 0.75) using a Perceiver-style cross-attention/self-attention stack; the inverter reconstructs both frames by decoding memory tokens back to patch space and unpatchifying.
Higher levels operate purely in token space, merging two token-latents into one and inverting one latent back into two.

\subsection{UCF-101 sweeper objective (local + path loss)}
Each sweeper is trained with (i) a local token-space reconstruction loss, and (ii) a path-sampled pixel reconstruction loss obtained by selecting a branch (left/right choices) and decoding through the frozen lower stack back to two frames. We also add a small merged-memory norm penalty.

\subsection{UCF-101 training schedule}
\begin{table}[h]
\centering
\begin{tabular}{lrrrr}
\toprule
Level & Videos & Epochs & LR & Pairs/video \\
\midrule
L0 & 1500 & 100 & $10^{-4}$ & -- \\
L1 & 4000 & 400 & $10^{-4}$ & 4 \\
L2 & 5000 & 320 & $10^{-4}$ & 4 \\
L3 & 5000 & 200 & $10^{-4}$ & 4 \\
\bottomrule
\end{tabular}
\caption{Hyperparameters for the level-wise training schedule}
\end{table}

\section{Baselines: Transformer and Mamba}
\label{app:baselines}

\subsection{Transformer sequence compressor}
Each frame is embedded with an MLP $\mathbb{R}^{784}\to\mathbb{R}^{d}$ (we use $d=1024$), a learnable \texttt{[CLS]} token is prepended, and learned positional embeddings are added. A Transformer encoder (6 layers, 8 heads, feedforward width $4d$, dropout 0.1) produces contextualized tokens, and the \texttt{[CLS]} output is used as a single compressed memory token. A Transformer decoder (6 layers, 8 heads) reconstructs the full sequence by cross-attending from learned per-timestep query vectors to the compressed token, followed by a per-frame MLP decoder $\mathbb{R}^{d}\to\mathbb{R}^{784}$.

\subsection{Mamba sequence compressor}
The Mamba baseline uses the same frame embedding MLP and the same Transformer decoder as above; only the sequence encoder is replaced by a Mamba backbone (12 layers, $d_{\mathrm{state}}=16$). We take the final timestep output as the compressed memory (kept as a single token for cross-attention).

\subsection{Baseline training protocol}
Unless stated otherwise, we train baselines for $T=128$ with reconstruction MSE, using AdamW (learning rate $10^{-4}$) with a cosine learning-rate schedule, batch size 16, gradient clipping at 1.0, and 200 epochs. We evaluate with per-pixel MSE and PSNR.

\section{Future Work}
\label{app:future_work}

\paragraph{Scaling to longer real-video horizons.}
Our real-video results are currently reliable only for 3--4 sweeper levels (16--32 frames), with the main failure mode being information loss in higher-level sweepers as latents are repeatedly merged. To push toward 128--512 frames without impractically large memory, we will (i) test more merge-friendly video representations (e.g., patch/grid or multi-scale latents that reduce high-frequency burden), and (ii) strengthen supervision at depth by replacing purely latent-level reconstruction with \emph{pixel-space} reconstruction at each level: after a merge/invert at level $\ell$, decode all the way back to images using the lower sweepers and image decoder and compute an image MSE against the original frames. This directly pressures high-level merges to preserve information that matters for end-to-end recovery. We will quantify gains by measuring how reconstruction error scales with depth on real video.

\paragraph{Audio and multimodal memory.}
We plan to extend IMFN to audio streams (waveforms or spectrogram/latent representations) to test whether the same local $2\!\rightarrow\!1$ merge primitive remains approximately invertible. We will then explore joint audio--video compression into a shared bounded memory state, leveraging cross-modal redundancy (e.g., audio cues disambiguating visually ambiguous events). Evaluation will measure reconstruction fidelity as well as robustness.

\paragraph{Student distillation for real-time robotics.}
IMFN's teacher supports efficient compression but still requires structured multi-stage computation; the distilled student provides constant-cost online updates suitable for streaming perception. We will evaluate students on long rollouts where frames arrive sequentially and the model is periodically queried for reconstructions or predictions, measuring retention under drift, latency/throughput, and stability over long horizons. Beyond reconstruction, we will test task utility in robotics-style settings (e.g., object permanence, tracking, event detection), and ultimately closed-loop deployment where the student memory conditions a lightweight policy, assessing whether bounded-memory compression remains robust when actions influence future observations.

\paragraph{Agentic AI and long-horizon planning with bounded memory.}
A natural application of IMFN is as a memory apparatus for agentic systems that must act over long horizons while operating under strict compute and storage budgets. We will investigate using the student as a constant-cost streaming memory that (i) maintains a compact state summarizing interaction history, and (ii) supports targeted \emph{retrieval-by-inversion}: given a query (e.g., ``where did I last see the red mug?'' or ``what changed since step $t$?''), the agent can invert a small portion of the hierarchy to reconstruct relevant segments for deliberation. We will benchmark agents on partially observable tasks requiring persistent state (e.g., navigation with revisitation, object permanence, instruction-following, and tool-use workflows where earlier outputs must be remembered). Concretely, we will compare (a) pure context-window baselines, (b) standard external-memory / key-value retrieval, and (c) IMFN-student memory with selective inversion, measuring task success, memory footprint, and query-time latency.

\end{document}